\documentclass[pdflatex,bst/sn-mathphys-num]{sn-jnl}

\usepackage{graphicx}%
\usepackage{multirow}%
\usepackage{amsmath,amssymb,amsfonts}%
\usepackage{amsthm}%
\usepackage{mathrsfs}%
\usepackage[title]{appendix}%
\usepackage{xcolor}%
\usepackage{textcomp}%
\usepackage{manyfoot}%
\usepackage{booktabs}%
\usepackage{algorithm}%
\usepackage{algorithmicx}%
\usepackage{algpseudocode}%
\usepackage{algpseudocode}%
\usepackage{algcompatible}%
\usepackage{tabularx}%
\usepackage{listings}%
\usepackage{natbib}%
\usepackage[compatibility=false]{caption}%

\begin{document}

\title[Article Title]{Unsupervised Latent Pattern Analysis for Estimating Type 2 Diabetes Risk in Undiagnosed Populations}

\author*[1]{\fnm{Praveen} \sur{Kumar}} 
\author[1]{\fnm{Vincent T.} \sur{Metzger}} 
\author[1]{\fnm{Scott A.} \sur{Malec}} 

\affil*[1]{\orgdiv{Department of Internal Medicine}, \orgname{University of New Mexico}, \city{Albuquerque}, \state{NM}, \country{USA}}


\abstract{
The global prevalence of diabetes, particularly type 2 diabetes mellitus (T2DM), is rapidly increasing, posing significant health and economic challenges. T2DM not only disrupts blood glucose regulation but also damages vital organs such as the heart, kidneys, eyes, nerves, and blood vessels, leading to substantial morbidity and mortality. In the US alone, the economic burden of diagnosed diabetes exceeded \$400 billion in 2022. Early detection of individuals at risk is critical to mitigating these impacts. While machine learning approaches for T2DM prediction are increasingly adopted, many rely on supervised learning, which is often limited by the lack of confirmed negative cases. To address this limitation, we propose a novel unsupervised framework that integrates Non-negative Matrix Factorization (NMF) with statistical techniques to identify individuals at risk of developing T2DM. Our method identifies latent patterns of multimorbidity and polypharmacy among diagnosed T2DM patients and applies these patterns to estimate the T2DM risk in undiagnosed individuals. By leveraging data-driven insights from comorbidity and medication usage, our approach provides an interpretable and scalable solution that can assist healthcare providers in implementing timely interventions, ultimately improving patient outcomes and potentially reducing the future health and economic burden of T2DM.
}


\maketitle

\section{Introduction}\label{sec1}

Diabetes is a chronic metabolic disorder that can lead to severe complications affecting the heart, blood vessels, eyes, kidneys, and nerves over time \citep{jbhi2025_1, jbhi2025_2}. Globally, its prevalence among adults aged $\geq$18 years increased from 7\% in 1990 to 14\% in 2022 \citep{jbhi2025_2}. The International Diabetes Federation projects that by 2050, about 853 million people will be living with diabetes worldwide \citep{jbhi2025_4}. Since 2000, global diabetes-related mortality has also risen. In 2021, diabetes caused 1.6 million deaths, contributed to about 530,000 kidney-related deaths, and was linked to 11\% of all cardiovascular deaths due to high blood glucose \citep{jbhi2025_2}.

In the US, diabetes is the $8^{th}$ leading cause of death \citep{jbhi2025_5}. In 2021, approximately 38.4 million Americans (11.6\% of the population), including 38.1 million adults (14.7\%) aged $\geq$18 years, were living with diagnosed or undiagnosed diabetes \citep{jbhi2025_6}. In 2022, the estimated economic burden of diagnosed diabetes totaled \$412.9 billion---\$306.6 billion in direct medical costs and \$106.3 billion in indirect costs \citep{jbhi2025_7}.

Given the rising prevalence, substantial morbidity and mortality, and significant economic burden of diabetes, early identification---particularly of undiagnosed individuals---is essential. With growing access to electronic health records (EHRs), claims data, and other real-world sources, researchers are increasingly applying various machine learning (ML) and statistical methods to predict or detect diabetes. To enhance predictive performance, \citep{jbhi2025_8} applied principal component analysis and minimum redundancy maximum relevance before training models such as Random Forests (RF) \citep{jbhi2025_12}, Neural Networks (NN) \citep{jbhi2025_26}, and Decision Trees (DT) \citep{jbhi2025_25}. For early detection of Type 2 Diabetes Mellitus (T2DM), \citep{jbhi2025_9} employed Gradient Boosting Machines (GBM) \citep{jbhi2025_27} and Logistic Regression (LR) \citep{jbhi2025_28} using demographic and laboratory data. Comparative analyses by \citep{jbhi2025_10} evaluated algorithms including LR, RF, Glmnet with LASSO \citep{jbhi2025_11}, XGBoost \citep{jbhi2025_13}, and LightGBM \citep{jbhi2025_14}. Other models---such as LR, RF, Gaussian Naïve Bayes (GNB), k-Nearest Neighbors (kNN) \citep{jbhi2025_16}, Classification and Regression Trees (CART) \citep{jbhi2025_17}, and Support Vector Machines (SVM) \citep{jbhi2025_18}---have also been applied to electronic medical records (EMRs) for T2DM risk prediction \citep{jbhi2025_15}. Population-level models have leveraged L1-regularized LR to identify surrogate variables for missing data \citep{jbhi2025_20}, and XGBoost has been used to predict T2DM incidence up to five years before clinical onset \citep{jbhi2025_21}. Traditional statistical methods---including the chi-square test, Student’s t-test, and Kruskal-Wallis test---have also been used for risk assessment in patients aged 45–75 without known diabetes \citep{jbhi2025_19}. Additionally, unsupervised learning methods such as UMAP \citep{jbhi2025_23} and DBSCAN \citep{jbhi2025_24} have been employed to explore socio-demographic clusters \citep{jbhi2025_22}.

These ML-based studies primarily focused on developing predictive models using binary classification techniques on datasets with labeled positive and negative examples. However, the lack of labeled negatives (a common issue in healthcare data where only positive cases are typically identified) can limit model performance and generalizability. To address this, we propose an unsupervised learning framework that algorithmically identifies latent patterns of comorbidities and pharmacotherapies among patients diagnosed with T2DM. We then use these data-driven patterns of multimorbidity and polypharmacy to estimate relative risk scores for T2DM in undiagnosed individuals. Our approach introduces methodological innovations to facilitate early detection and prioritize high-risk individuals for preventive intervention.

\section{Materials and Methods}

This study used claims data to develop an interpretable, scalable unsupervised framework for estimating T2DM risk. This section describes the data source, cohort development, phenotyping methods, covariate selection, proposed algorithms, and validation approaches.

\subsection{Data Source and Cohort Development}
We used data from the US Merative MarketScan Commercial Claims and Encounters (CCAE) database (2018–2022), standardized using the Observational Medical Outcomes Partnership Common Data Model version 5 (OMOP CDMv5) \citep{jbhi2025_29}. As a proof-of-concept study aimed at developing a generalizable methodology to estimate the risk of T2DM, the CCAE database was an appropriate choice due to its large sample size and comprehensive claims coverage. To define the study cohort, individuals were required to have at least 730 days of continuous observation starting on or after January 1, 2018. Although the official transition from the International Classification of Diseases, Ninth Revision, Clinical Modification (ICD-9-CM) to ICD-10-CM occurred on October 1, 2015, the database still contained many patients with ICD-9-CM codes in 2016 and 2017. To ensure consistency and uniformity in diagnostic coding, we restricted our analysis to ICD-10-CM diagnosis codes. Accordingly, individuals whose observation began before January 1, 2018, were excluded. Applying these criteria, we identified a cohort of 17,118,687 individuals, comprising both diagnosed and undiagnosed cases.

\subsection{Phenotyping and Covariate Selection}
Patients diagnosed with T2DM were identified by the presence of ICD-10-CM code \textbf{E11.*} in their medical records; all others were considered undiagnosed (unlabeled). For the machine learning (ML) models, we extracted 66,716 unique covariates from both diagnosed and undiagnosed patients. The list of covariates excluded any E11.* codes. Diagnosed individuals (1,338,467; 7.82\%) had covariates recorded between the start of their observation period and the day before their first T2DM diagnosis, while undiagnosed individuals (15,780,220; 92.18\%) had covariates recorded over their entire observation period. This strategy aimed to capture covariates associated with the T2DM development, assuming that  undiagnose individuals with similar covariate profiles may also be at risk. Covariates included clinical diagnoses (ICD-10-CM codes) and prescribed medications (RxNorm codes). A compressed sparse row (CSR) matrix of size $17,118,687 \times 66,716$ was created to represent all patients and their covariates. For each patient, absent covariates were assigned a value of 0, and present covariates were assigned the number of encounters during the selected period.

\subsection{Machine Learning Methods and Proposed Algorithms}

Claims datasets typically contain only confirmed positives (i.e., diagnosed individuals) and lack confirmed negatives for conditions such as T2DM. Importantly, the absence of a diagnosis does not imply the absence of disease; many individuals in the undiagnosed group may, in fact, have T2DM but remain unrecorded due to underdiagnosis or incomplete documentation. As a result, the undiagnosed population likely comprises a mixture of true negatives and unlabeled positives. Treating all undiagnosed individuals as true negatives in traditional supervised binary classification can introduce substantial label noise and bias, ultimately degrading model performance \citep{jbhi2025_33}. In such settings, positive and unlabeled (PU) learning provides a more appropriate framework for learning from data containing only positive labels and an unlabeled set \citep{jbhi2025_31, jbhi2025_32}.

\begin{figure}[h]
\centering
\includegraphics[width=0.99\columnwidth]{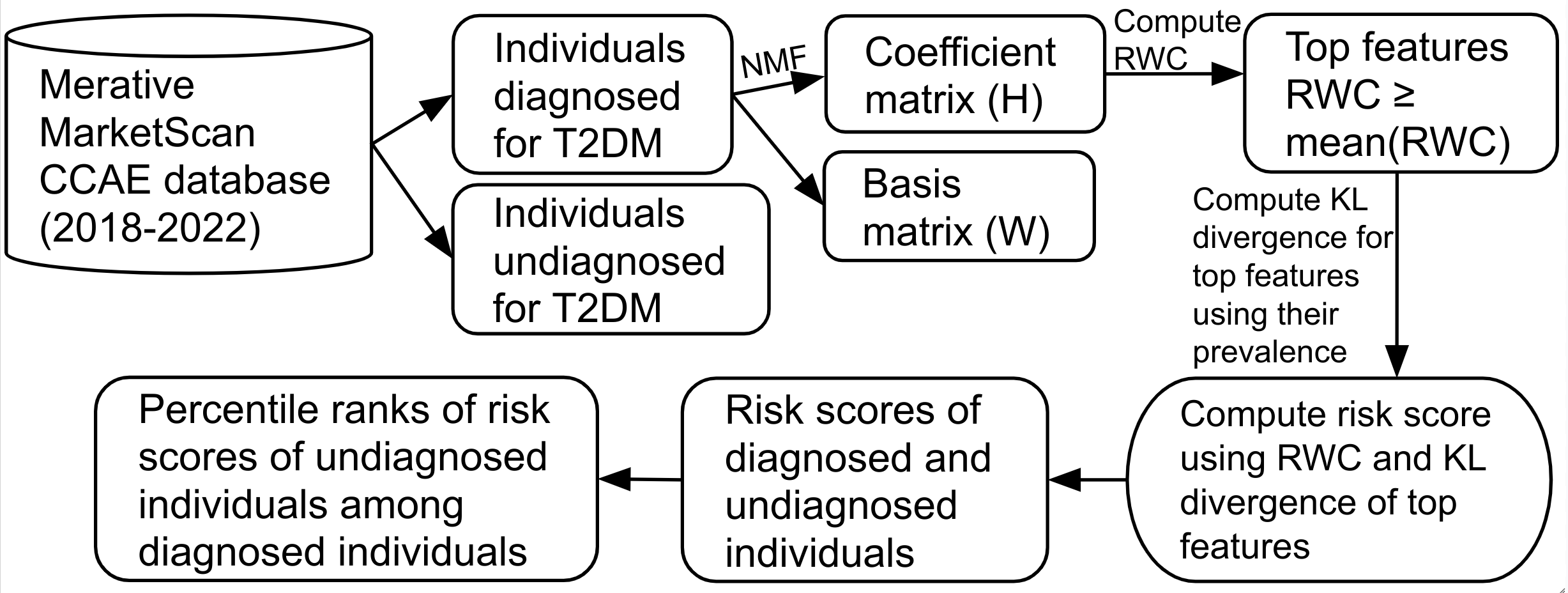}
\caption{\textbf{Steps to estimate T2DM risk score}. RWC: Rank-weighted coefficients of covariates. KL divergence: Kullback-Leibler divergence.}
\label{fig:NMF_steps}
\end{figure}

We did not employ PU learning in this study. Instead, we developed a novel unsupervised framework that uses only labeled positive examples to identify individuals at risk of T2DM within the undiagnosed population. Figure \ref{fig:NMF_steps} outlines the methodology: 1) apply Non-negative Matrix Factorization (NMF) \citep{jbhi2025_34} to the data of diagnosed individuals; 2) compute the Rank-Weighted Coefficients (RWC) for all covariates using the coefficient matrix from NMF; 3) remove covariates with RWC = 0 and select those with RWC $\geq$ mean(RWC); 4) determine the prevalence of each selected covariate in both diagnosed and undiagnosed groups; 5) compute the Kullback–Leibler (KL) divergence \citep{jbhi2025_35} for the selected covariates based on their prevalences; 6) use KL divergence and RWC values of the selected covariates to estimate risk scores for diagnosed and undiagnosed individuals; and 7) compute the percentile rank of each undiagnosed individual's risk score relative to the distribution of risk scores among diagnosed individuals. The following subsections provide detailed descriptions of each step.

\subsection{Training and Validation Sets}

Although the CSR matrix initially included 66,716 covariates for each individual, diagnosed patients had an average of only 18 non-zero covariates, with some having as few as one or two non-zero covariates. To ensure the robustness of the NMF models, we included only those diagnosed patients with at least five non-zero covariates, resulting in a final matrix of size $810,040 \times 66,716$. From these 810,040 diagnosed patients, 8,100 were randomly selected as validation records, and the remaining 801,940 were used to train the NMF models.

\subsection{Non-negative Matrix Factorization (NMF) for Feature Selection}

Non-negative Matrix Factorization (NMF) is an unsupervised dimensionality reduction technique that identifies latent patterns in high-dimensional, non-negative data. Given a non-negative matrix $X \in \mathbb{R}^{n \times m}$, where $n$ is the number of samples and $m$ is the number of features (e.g., diagnoses, medications), NMF factorizes $X$ into two lower-rank non-negative matrices: a basis matrix $W \in \mathbb{R}^{n \times k}$ and a coefficient matrix $H \in \mathbb{R}^{k \times m}$, such that: $X \approx WH$. Here, $k \ll \min(n, m)$ denotes the number of latent components.

The coefficient matrix $H$ captures the contribution of each feature to each latent component (rows of $H$ correspond to components and columns to features). The entry $H_{ij}$ in $H$ matrix represents the contribution of the $j$-th feature to the $i$-th latent component. If a feature $j$ has consistently high contribution, $H_{ij}$, across multiple rows of $H$, it suggests that the feature is important for reconstructing multiple latent components and is thus highly relevant to the underlying structure of the data. To quantify the importance of each feature, we define a Rank-Weighted Coefficient (RWC), $w_j$, which incorporates both the magnitude of a feature's contribution, $H_{ij}$, and its rank, $r_{ij}$, within each component $i$. Here, rank refers to a feature's position based on its contribution magnitude, $H_{ij}$, within a component, where a smaller numerical rank (e.g., 1st place) indicates a larger contribution than a higher rank (e.g., 10th place). To emphasize features with consistently high contributions across components, we used the reciprocal of the rank in computing RWC. The RWC for a feature $j$ is calculated as follows (Equation \ref{eq:rwc_formula}):

\begin{equation}
\text{w}_j = \dfrac{ \sum_{i=1}^{k} H_{ij} \times \dfrac{1}{r_{ij}} }{ \sum_{i=1}^{k} \dfrac{1}{r_{ij}} }
\label{eq:rwc_formula}
\end{equation}

We ran NMF models on the training set comprising 801,940 diagnosed individuals and 66,716 covariates (features), represented as an $801,940 \times 66,716$ CSR matrix. The models were executed for 40 iterations using different random seeds, and the RWC was computed for each feature in every iteration. The final RWC for each feature was then calculated by averaging the RWC values across all 40 iterations. We manually reviewed the features and their corresponding RWC scores and found that only 38,234 features (57.31\%) had a non-zero RWC, with the highest RWC value being 98.22 and the lowest $1.05 \times 10^{-10}$. Additionally, we observed a sharp decline in RWC values among lower-ranked features. To refine the feature set, we computed the mean RWC using all features with non-zero RWC and selected only those with an RWC $\geq \text{mean(RWC)}$ for the subsequent analysis. The mean RWC was 0.0384, and 1,629 features exceeded this threshold. Thus, by applying NMF exclusively to the data of diagnosed patients, we selected 1,629 co-occurring conditions and medications commonly present in the diagnosed population.

\subsubsection{Determine the number of components (clusters) for NMF}

The NMF algorithm requires the number of components ($k$) to be specified in advance. To determine the optimal value of $k$, we executed NMF for $k \in [2, 15]$ on the training data, represented as an $801,940 \times 66,716$ CSR matrix. For each value of $k$, we computed the reconstruction error based on the Frobenius norm (Equation \ref{eq:fnorm_formula}). By plotting the reconstruction error against $k$, we identified an ``elbow'' point---where the marginal improvement in error reduction began to plateau. Using this approach, we selected $k = 9$ as the number of components for our NMF models. Notably, this choice aligns with the nine major subcategories in ICD-10-CM code `E11.*' for T2DM \citep{jbhi2025_30}.

\begin{equation}
\|X - WH\|_F = \sqrt{\sum_{i=1}^{n}\sum_{j=1}^{m} (x_{ij} - (WH)_{ij})^2}
\label{eq:fnorm_formula}
\end{equation}
where $X \in \mathbb{R}^{n \times m}$ is the original matrix and $WH$ is its NMF approximation.

\subsection{Feature Prevalence in Diagnosed and Undiagnosed Populations}

We define the \textit{feature prevalence} of covariate $j$ as the proportion of individuals in a given group (diagnosed or undiagnosed) who have at least one recorded instance of that covariate. Let $n_d$ and $n_u$ denote the number of diagnosed and undiagnosed individuals, respectively. The prevalence of feature $j$ in the diagnosed and undiagnosed populations, denoted as $p_j^{(d)}$ and $p_j^{(u)}$, is computed as follows (Equations \ref{eq:diag_prev} and \ref{eq:undiag_prev}):
\begin{equation}
p_j^{(d)} = \frac{1}{n_d} \sum_{i=1}^{n_d} \mathbb{I}(x_{ij} > 0)
\label{eq:diag_prev}
\end{equation}

\begin{equation}
p_j^{(u)} = \frac{1}{n_u} \sum_{i=1}^{n_u} \mathbb{I}(x_{ij} > 0)
\label{eq:undiag_prev}
\end{equation}

where $x_{ij}$ represents the value of feature $j$ for individual $i$, and $\mathbb{I}(x_{ij} > 0)$ is the indicator function defined as follows (Equation \ref{eq:indicator_function}):

\begin{equation}
\mathbb{I}(x_{ij} > 0) =
\begin{cases}
1 & \text{if } x_{ij} > 0 \\
0 & \text{otherwise}
\end{cases}
\label{eq:indicator_function}
\end{equation}

Using Equations \ref{eq:diag_prev} and \ref{eq:undiag_prev}, we computed the \textit{feature prevalence} of all 1,629 NMF-selected covariates in the diagnosed and undiagnosed groups, respectively.

\subsection{Kullback–Leibler (KL) Divergence for Covariates}

The Kullback--Leibler (KL) divergence measures the relative entropy between two probability distributions, quantifying how one distribution diverges from another. To assess the discriminative power of each feature between diagnosed and undiagnosed populations, we computed the KL divergence, $\text{d}_j$, for each feature $j$ as follows (Equation \ref{eq:kld_formula}):

\begin{equation}
\begin{aligned}
d_j &= D_{\mathrm{KL}}\left( p_j^{(d)} \,\middle\|\, p_j^{(u)} \right) \\
    &= \left(p_j^{(d)} + \varepsilon\right) \log\left( \frac{p_j^{(d)} + \varepsilon}{p_j^{(u)} + \varepsilon} \right) \\
    &\quad + \left(1 - p_j^{(d)} + \varepsilon\right) \log\left( \frac{1 - p_j^{(d)} + \varepsilon}{1 - p_j^{(u)} + \varepsilon} \right)
\end{aligned}
\label{eq:kld_formula}
\end{equation}

where $p_j^{(d)}$ (Equation \ref{eq:diag_prev}) and $p_j^{(u)}$ (Equation \ref{eq:undiag_prev}) represent the prevalence of feature $j$ in the diagnosed and undiagnosed populations, respectively. The small constant $\varepsilon$ (e.g., $10^{-8}$) is added to avoid issues with $\log(0)$ or divide by zero. Using Equation \ref{eq:kld_formula}, we computed the KL divergence ($\text{d}_j$) for all 1,629 NMF-selected covariates.

A KL divergence ($\text{d}_j$) of zero indicates that feature $j$ has identical prevalence in both the diagnosed and undiagnosed populations and can therefore be considered uninformative. A large KL divergence suggests that feature $j$ is more prevalent among diagnosed individuals, indicating potential clinical relevance to the target condition.

\subsection{Risk Estimation using KL Divergence and NMF-derived RWC}

We used the rank-weighted coefficients ($w_j$) and KL divergence values ($d_j$) of the 1,629 covariates selected by NMF (denoted as the set $\mathcal{S}$) to compute the T2DM risk score ($r_i$) for each diagnosed and undiagnosed patient as follows (Equation \ref{eq:risk_score}):

\begin{equation}
r_i = \sum_{j \in \mathcal{S}} w_j \cdot d_j \cdot \mathbb{I}(x_{ij} > 0)
\label{eq:risk_score}
\end{equation}

Here, $x_{ij}$ represents the value of covariate $j$ for individual $i$, and $\mathbb{I}(x_{ij} > 0)$ is the indicator function. To convert the raw risk score $r_i$ to the range $[0, 1)$, we applied a scaled arctangent transformation (Equation \ref{eq:arctan_prob}). The resulting score $\tilde{r}_i \in [0, 1)$ provides improved interpretability; however, these transformed scores should not be interpreted as calibrated probabilities. While the sigmoid function is a commonly used alternative for such purposes, it exhibits rapid asymptotic behavior—approaching its upper limit of 1 quickly as the input score value increases. This property makes it highly sensitive to outliers and can result in overconfident estimates ($\sim1$). In contrast, the arctangent function increases more gradually, reducing the influence of extreme scores and providing more stable estimates. This characteristic makes it particularly suitable in scenarios where the underlying score distribution does not follow a logistic pattern.

\begin{equation}
\tilde{r}_i = \frac{2}{\pi} \tan^{-1}(r_i)
\label{eq:arctan_prob}
\end{equation}

Finally, we computed the percentile rank of the risk score for each undiagnosed individual with respect to the distribution of risk scores among diagnosed patients. Based on these percentiles, we categorized the risk scores of undiagnosed individuals into three broader risk categories: 
\textit{low} ($<50^{th}$ percentile), 
\textit{moderate} ($50^{th}-90^{th}$ percentile), and 
\textit{high} ($>90^{th}$ percentile). 
These category thresholds were determined in consultation with a subject matter expert (SME), who reviewed the distribution of risk scores and the associated covariates among diagnosed individuals.

\subsection{Algorithms}
Algorithm \ref{alg:RiskScoreEstimator} presents the pseudocode for computing the risk score for each diagnosed and undiagnosed patient. Algorithms \ref{alg:RWCEstimator} and \ref{alg:KLDEstimator} are subroutines used to calculate the RWC and KL divergence, respectively. 

\begin{algorithm}
    \caption{T2DM Risk Score Estimation}
    \label{alg:RiskScoreEstimator}
    \textbf{Input}: $X = X_d \cup X_u$, covariates $\mathcal{F}$, number of components $k$ \\ 
    \Comment{$X_d$: diagnosed patient matrix, $X_u$: undiagnosed patient matrix} \\
    \textbf{Output}: Estimated risk scores $\tilde{r}_i$ for each patient $i$
    \begin{algorithmic}[1]
        \STATE $w \gets \text{compute\_rwc}(X_d, \mathcal{F}, k)$ \Comment{Rank-weighted coefficients (RWC)}
        \STATE $\mathcal{S} \gets \{j \in \mathcal{F} \mid w_j \geq \text{mean}(w)\}$ \Comment{Select covariates with RWC $\geq$ mean}
        \STATE $\hat{X}_d \gets X_d[\mathcal{S}]$ \Comment{Diagnosed patient matrix with selected covariates}
        \STATE $\hat{X}_u \gets X_u[\mathcal{S}]$ \Comment{Undiagnosed patient matrix with selected covariates}
        \STATE $d \gets \text{compute\_kld}(\hat{X}_d, \hat{X}_u, \mathcal{S})$ \Comment{KL divergence values}
        \STATE $\hat{X} \gets \hat{X}_d \cup \hat{X}_u$ \Comment{Full matrix with selected covariates}
        \FOR{each patient $i = 1$ to $|\hat{X}|$}
            \STATE $r_i \gets \sum_{j \in \mathcal{S}} w_j \cdot d_j \cdot \mathbb{I}(x_{ij} > 0)$ \Comment{Raw risk score}
            \STATE $\tilde{r}_i \gets \frac{2}{\pi} \cdot \arctan(r_i)$ \Comment{Normalized risk score in [0, 1)}
        \ENDFOR
        \STATE \textbf{return} $\tilde{r}_i$ for all $i$
    \end{algorithmic}
\end{algorithm}

\begin{algorithm}
    \caption{Compute Rank-Weighted Coefficients (RWC) for Covariates (\texttt{compute\_rwc()})}
    \label{alg:RWCEstimator}
    \textbf{Input}: $X_d$, covariates $\mathcal{F}$, number of components $k$ \\
    \textbf{Output}: $w_j$ for $j \in \mathcal{F}$ \Comment{RWC values}
    \begin{algorithmic}[1]
        \STATE $W, H \gets \text{NMF}(X_d, k)$ \Comment{$H \in \mathbb{R}^{k \times |\mathcal{F}|}$}
        \FOR{each $j \in \mathcal{F}$}
            \STATE $num \gets 0$ \Comment{Initialize numerator}
            \STATE $den \gets 0$ \Comment{Initialize denominator}
            \FOR{$i = 1$ to $k$} \Comment{Loop over components}
                \STATE $r_{ij} \gets$ rank of covariate $j$ among all covariates in component $i$, based on descending $H_{ij}$
                \STATE $num \gets num + H_{ij} \cdot \frac{1}{r_{ij}}$
                \STATE $den \gets den + \frac{1}{r_{ij}}$ 
            \ENDFOR
            \STATE $w_j \gets \frac{num}{den}$ \Comment{Compute RWC for covariate $j$}
        \ENDFOR
        \STATE \textbf{return} $\{w_j\}_{j \in \mathcal{F},\, w_j > 0}$
    \end{algorithmic}
\end{algorithm}

\begin{algorithm}
    \caption{KL Divergence Computation \\ (\texttt{compute\_kld}())}
    \label{alg:KLDEstimator}
    \textbf{Input}: $\hat{X}_d$, $\hat{X}_u$, selected covariates $\mathcal{S}$ \\
    \textbf{Output}: $d_j$ for all $j \in \mathcal{S}$ \Comment{KL divergence values}
    \begin{algorithmic}[1]
        \STATE $\epsilon \gets 10^{-8}$ 
        \FOR{each $j \in \mathcal{S}$}
            \STATE $p_j^{(d)} \gets \frac{1}{|\hat{X}_d|} \sum_{i=1}^{|\hat{X}_d|} \mathbb{I}(\hat{x}_{d_{ij}} > 0)$ \Comment{ $\hat{x}_{d_{ij}}$: value of the $j$-th covariate for patient $i$ in $\hat{X}_d$}
            \STATE $p_j^{(u)} \gets \frac{1}{|\hat{X}_u|} \sum_{i=1}^{|\hat{X}_u|} \mathbb{I}(\hat{x}_{u_{ij}} > 0)$ \Comment{ $\hat{x}_{u_{ij}}$: value of the $j$-th covariate for patient $i$ in $\hat{X}_u$}
            \STATE $d_j \gets (p_j^{(d)} + \epsilon) \cdot \log \left( \frac{p_j^{(d)} + \epsilon}{p_j^{(u)} + \epsilon} \right)$
            \STATE \hspace{1.5em} $+ (1 - p_j^{(d)} + \epsilon) \cdot \log \left( \frac{1 - p_j^{(d)} + \epsilon}{1 - p_j^{(u)} + \epsilon} \right)$
        \ENDFOR
        \STATE \textbf{return} $\{d_j\}_{j \in \mathcal{S}}$
    \end{algorithmic}
\end{algorithm}

\subsection{Validation Methods}
We used the following four approaches to validate the results and effectiveness of our proposed method:
\subsubsection{Method 1}
We computed the mean, median, and maximum predicted risk scores separately for individuals in the training, validation, and undiagnosed cohorts. Since both the training and validation sets comprised diagnosed individuals, we expected their risk score distributions to be similar and substantially higher than those of the undiagnosed cohort. This comparison serves as a form of temporal validation, evaluating the model's ability to identify individuals at elevated risk who subsequently received a clinical diagnosis of T2DM, thereby simulating a real-world early detection scenario.

\subsubsection{Method 2}
To assess the clinical relevance of the model’s predicted T2DM risk scores, we randomly selected 100 undiagnosed individuals from each percentile bin (1–100) based on their predicted risk scores. Three AI models (Llama, ChatGPT, and Perplexity) and SME were presented with the covariates for each individual and asked to classify their T2DM risk into one of three categories: \textit{low}, \textit{moderate}, and \textit{high}. Both the SME and the AI models were blinded to the individuals' percentile ranks to prevent bias. We then evaluated the level of agreement between our model’s risk category assignments and those determined by the SME and AI models across the three risk strata. Disagreements among AI models were resolved via majority voting.

\subsubsection{Method 3}
To evaluate the robustness of the model’s risk estimation, we hypothesized that high-risk diagnosed individuals and high-risk undiagnosed individuals would share similar covariate patterns, while high-risk diagnosed individuals and low-risk undiagnosed individuals would show dissimilar profiles. To test this, we identified all diagnosed and undiagnosed individuals with predicted risk scores $\geq$0.90 and computed the Jaccard similarity between the top 10 to 500 most frequent covariates in each group. We performed a similar analysis by comparing diagnosed individuals with risk scores $\geq$0.90 to undiagnosed individuals with risk scores $\leq$0.10, again using the Jaccard similarity across the top 10 to 500 covariates.

\subsubsection{Method 4}
To evaluate the discriminative power of our model and its ability to identify probable T2DM cases within the undiagnosed population, we conducted the following experiment. All diagnosed individuals were treated as positive cases, and an equal number of undiagnosed individuals were selected as negative examples based on three different risk score thresholds ($\leq$0.1, $\leq$0.5, and $\leq$1.0). We then trained and evaluated XGBoost classifiers on these balanced datasets using 5-fold cross-validation over 40 iterations and calculated classification metrics with 95\% confidence intervals (CI). A balanced design was used to eliminate the influence of class imbalance, ensuring that any observed performance variation was attributable to differences in label quality rather than sample distribution. We hypothesized that classifiers trained on datasets where negative examples were drawn from the low-risk ($\leq$0.1) undiagnosed group would exhibit superior performance, as these individuals are more likely to be true negatives. In contrast, classifiers trained using higher-risk undiagnosed individuals as negatives ($\leq$0.5, and $\leq$1.0) are expected to perform worse due to label noise since these groups may include a mixture of true negatives and undiagnosed true positives.

\section{Results}\label{sec2}

Applying our inclusion and exclusion criteria, we identified 17,118,687 individuals (8,249,175 males and 8,869,512 females) for our study. Of these, 1,338,467 (7.82\%) had a diagnosis of T2DM, while the remaining 15,780,220 (92.18\%) did not have a coded diagnosis of T2DM. Table \ref{tab:patient_characteristics} provides a comprehensive summary of patient characteristics. We noticed a higher proportion of males were coded for T2DM (51.68\% vs. 48.32\%). Patients diagnosed with T2DM were generally older, with 42.87\% aged 50–59 and 24.17\% aged 40–49.

\begin{table*}[h]
\caption{
\textbf{Characteristics of diagnosed and undiagnosed patients.} \textnormal{Comorbidities and medications are top covariates with high RWC values.}}
\makebox[\textwidth][c]{%
\resizebox{\textwidth}{!}{%
\begin{tabular}{|l|l|l|}
\hline
\textbf{\begin{tabular}[c]{@{}l@{}}Patient Characteristics\\ (n=17,118,687)\end{tabular}} & \textbf{\begin{tabular}[c]{@{}l@{}}Diagnosed\\ (n=1,338,467)\end{tabular}} & \textbf{\begin{tabular}[c]{@{}l@{}}Undiagnosed\\ (n=15,780,220)\end{tabular}} \\ \hline
Male & 691,676 (51.68\%) & 7,557,499 (47.89\%) \\ \hline
Female & 646,791 (48.32\%) & 8,222,721 (52.11\%) \\ \hline
\textbf{Age} &  &  \\ \hline
0-9 & 1,770 (0.13\%) & 2,194,145 (13.90\%) \\ \hline
10-19 & 12,510 (0.93\%) & 2,585,382 (16.38\%) \\ \hline
20-29 & 42,454 (3.17\%) & 2,397,720 (15.19\%) \\ \hline
30-39 & 130,851 (9.78\%) & 2,574,401 (16.31\%) \\ \hline
40-49 & 323,495 (24.17\%) & 2,556,859 (16.20\%) \\ \hline
50-59 & 573,774 (42.87\%) & 2,564,958 (16.25\%) \\ \hline
$\geq$60 & 253,613 (18.95\%) & 906,755 (5.75\%) \\ \hline
\textbf{Comorbidities} &  &  \\ \hline
End Stage Renal Disease & 3,931 (0.29\%) & 10,944 (0.07\%) \\ \hline
\begin{tabular}[c]{@{}l@{}}Secondary Hyperparathyroidism of Renal Origin\end{tabular} & 3,174 (0.24\%) & 13,879 (0.09\%) \\ \hline
Essential (Primary) Hypertension & 361,479 (27.01\%) & 2,782,280 (17.63\%) \\ \hline
Iron Deficiency Anemia & 28,773 (2.15\%) & 434,137 (2.75\%) \\ \hline
Obstructive Sleep Apnea & 129,031 (9.64\%) & 927,077 (5.87\%) \\ \hline
Cervicalgia & 85,187 (6.36\%) & 1,765,717 (11.19\%) \\ \hline
Hyperlipidemia & 192,259 (14.36\%) & 2,125,971 (13.47\%) \\ \hline
Hypothyroidism & 91,513 (6.84\%) & 1,056,640 (6.70\%) \\ \hline
Obesity & 130,239 (9.73\%) & 1,544,672 (9.79\%) \\ \hline
Hydrochlorothiazide & 147,401 (11.01\%) & 866,481 (5.49\%) \\ \hline
\end{tabular}
}
}
\label{tab:patient_characteristics}
\end{table*}

According to our method's estimates, out of 15,780,220 undiagnosed individuals, 502,144 (3.18\%) had T2DM risk scores that fell above the 90th percentile (\textit{High risk}) of the diagnosed individuals' risk scores, 3,178,038 (20.14\%) fell in the 50th to 90th percentile (\textit{Moderate risk}), and 12,100,038 (76.68\%) fell below the 50th percentile (\textit{Low risk}). The percentage of undiagnosed individuals in each risk category across age groups is shown in Table \ref{tab:dist_by_age_group}.

\begin{table*}[h]
\caption{
\textbf{Percentage of undiagnosed individuals in each risk category by age group.} \textnormal{Total Count: total undiagnosed individuals in age group.}}
\makebox[\textwidth][c]{%
\begin{tabular}{|l|l|l|l|l|}
\hline
\textbf{Age } & \textbf{Low } & \textbf{Moderate } & \textbf{High } & \textbf{Total Count } \\ \hline
0-9 & 97.95\% & 2.02\% & 0.03\% & 2,194,145 \\ \hline
10-19 & 96.10\% & 3.80\% & 0.10\% & 2,585,382 \\ \hline
20-29 & 90.67\% & 8.73\% & 0.60\% & 2,397,720 \\ \hline
30-39 & 79.38\% & 18.09\% & 2.53\% & 2,574,401 \\ \hline
40-49 & 64.59\% & 29.51\% & 5.90\% & 2,556,859 \\ \hline
50-59 & 50.76\% & 40.53\% & 8.71\% & 2,564,958 \\ \hline
$\geq$60 & 43.42\% & 47.39\% & 9.18\% & 906,755 \\ \hline
\end{tabular}
}
\label{tab:dist_by_age_group}
\end{table*}

\subsection{Characterization of T2DM Pattern by State }
\begin{figure}[H]
\centering
\includegraphics[width=0.97\columnwidth]{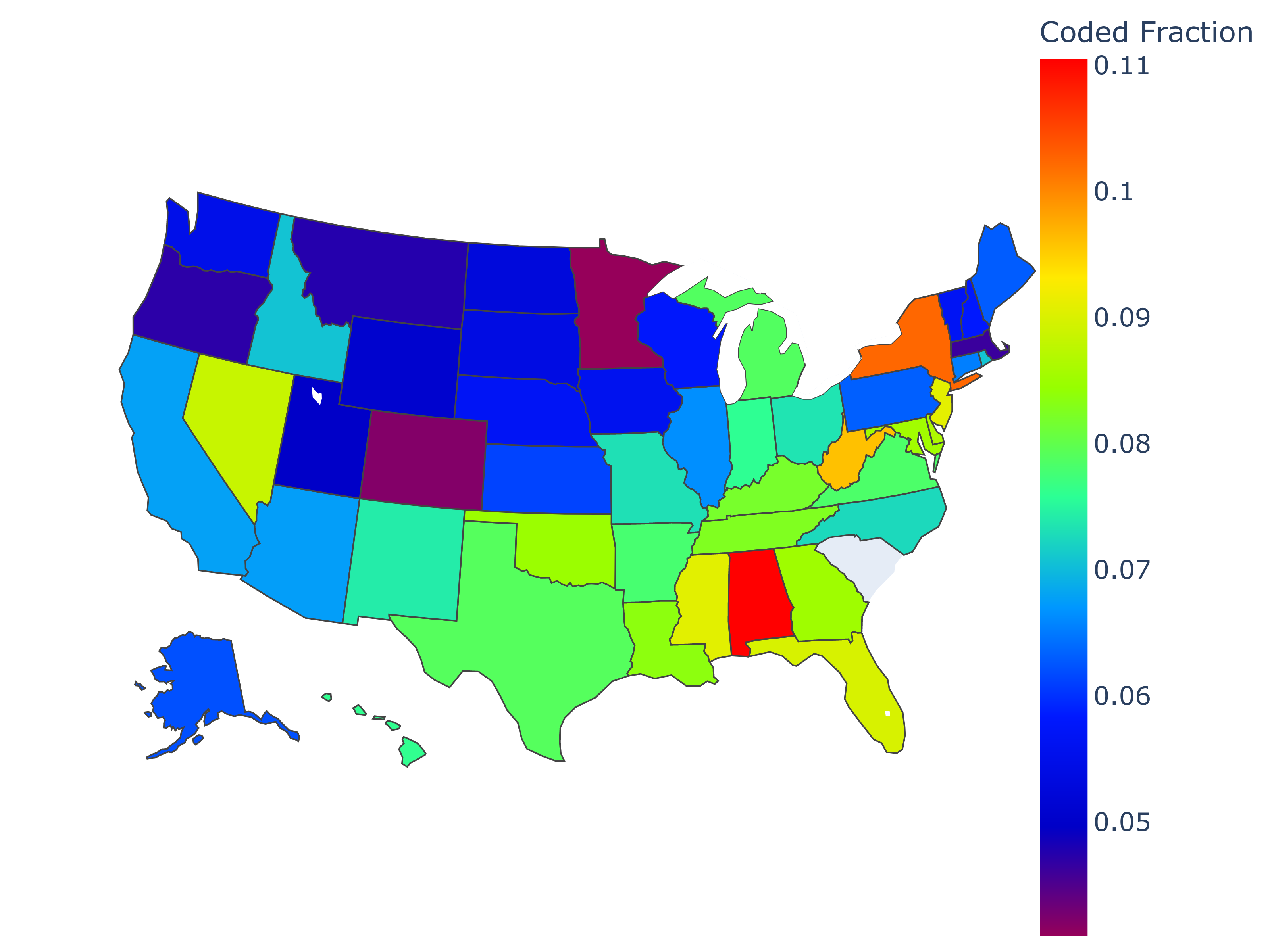}
\caption{\textbf{T2DM coded fraction by US state.} South Carolina was excluded due to license restrictions; Puerto Rico was excluded due to having a small sample size (n=109).}
\label{fig:us_state_plot}
\end{figure}

The coded fraction of T2DM, defined as the proportion of explicitly coded cases relative to the total sample size within each U.S. state, is illustrated in Figure \ref{fig:us_state_plot}. The coded fraction ranged from 4.10\% in Minnesota (MN) to 11.05\% in Alabama (AL).

\subsection{Validation Method 1}

Table \ref{tab:validation_method_1} presents the mean, median, and maximum predicted T2DM risk scores for the training, validation, and undiagnosed patient cohorts. As anticipated, the training and validation cohorts—both comprising diagnosed individuals—have comparable risk score distributions across all three summary statistics. In contrast, the undiagnosed cohort shows substantially lower mean and median risk scores, indicating a clear distinction from the diagnosed population. However, the maximum risk score in the undiagnosed group is comparable to that of the diagnosed cohorts, suggesting the presence of potentially high-risk individuals who may require further clinical evaluation.

\begin{table*}[h]
\caption{
\textbf{Predicted mean, median, and maximum T2DM risk scores across cohorts. }\textnormal{Training and validation sets (diagnosed) had only pre-diagnosis covariates, while the undiagnosed set included all covariates from the observation period.}}
\makebox[\textwidth][c]{%
\begin{tabular}{|l|l|l|l|}
\hline
\textbf{Data} & \textbf{Mean} & \textbf{Median} & \textbf{Maximum} \\ \hline
Training set (n=801,940) & 0.5577 & 0.5262 & 0.9679 \\ \hline
Validation set (n=8,100) & 0.5651 & 0.5667 & 0.9671 \\ \hline
undiagnosed set (n=15,780,220) & 0.3310 & 0.1743 & 0.9681 \\ \hline
\end{tabular}
}
\label{tab:validation_method_1}
\end{table*}

\subsection{Validation Method 2}

Table \ref{tab:validation_method_2} presents our model-assigned risk categories alongside classifications made by the SME/AI models for 100 randomly selected undiagnosed individuals. The overall agreement between our model and AI-based assessments was 89\%, while agreement with SME assessments was 75\%, indicating substantial consistency with both clinical judgment and AI-supported reasoning. Both AI (85.71\%) and SME (89.80\%) showed higher agreement with the model for individuals classified as low risk. The AI showed its highest agreement (95.12\%) for individuals in the moderate-risk category, while SME had their highest agreement (89.80\%) in the low-risk category. Most discrepancies occurred near category boundaries, which is expected given the subjective nature of risk interpretation in these transitional ranges.

\begin{table*}[h]
\caption{
\textbf{Agreement between model-predicted and SME/AI risk classifications for 100 undiagnosed individuals.} \textnormal{The ``Agreement'' column reflects the proportion where SME/AI matches the model's classification.}}
\makebox[\textwidth][c]{%
\begin{tabular}{|l|lll|l|l|}
\hline
 & \multicolumn{3}{c|}{\textbf{AI Classification}} &  &  \\ \hline
\textbf{Model classification} & \multicolumn{1}{l|}{Low} & \multicolumn{1}{l|}{Moderate} & High & Total Count & Agreement \\ \hline
Low & \multicolumn{1}{l|}{\textbf{42}} & \multicolumn{1}{l|}{7} & 0 & 49 & 85.71\% \\ \hline
Moderate & \multicolumn{1}{l|}{0} & \multicolumn{1}{l|}{\textbf{39}} & 2 & 41 & 95.12\% \\ \hline
High & \multicolumn{1}{l|}{0} & \multicolumn{1}{l|}{2} & \textbf{8} & 10 & 80.00\% \\ \hline
 & \multicolumn{3}{c|}{\textbf{SME Classification}} &  &  \\ \hline
Low & \multicolumn{1}{l|}{\textbf{44}} & \multicolumn{1}{l|}{5} & 0 & 49 & 89.80\% \\ \hline
Moderate & \multicolumn{1}{l|}{12} & \multicolumn{1}{l|}{\textbf{25}} & 4 & 41 & 60.98\% \\ \hline
High & \multicolumn{1}{l|}{0} & \multicolumn{1}{l|}{4} & \textbf{6} & 10 & 60.00\% \\ \hline
\end{tabular}
}
\label{tab:validation_method_2}
\end{table*}

\subsection{Validation Method 3}
\begin{figure}[H]
\centering
\includegraphics[width=0.97\columnwidth]{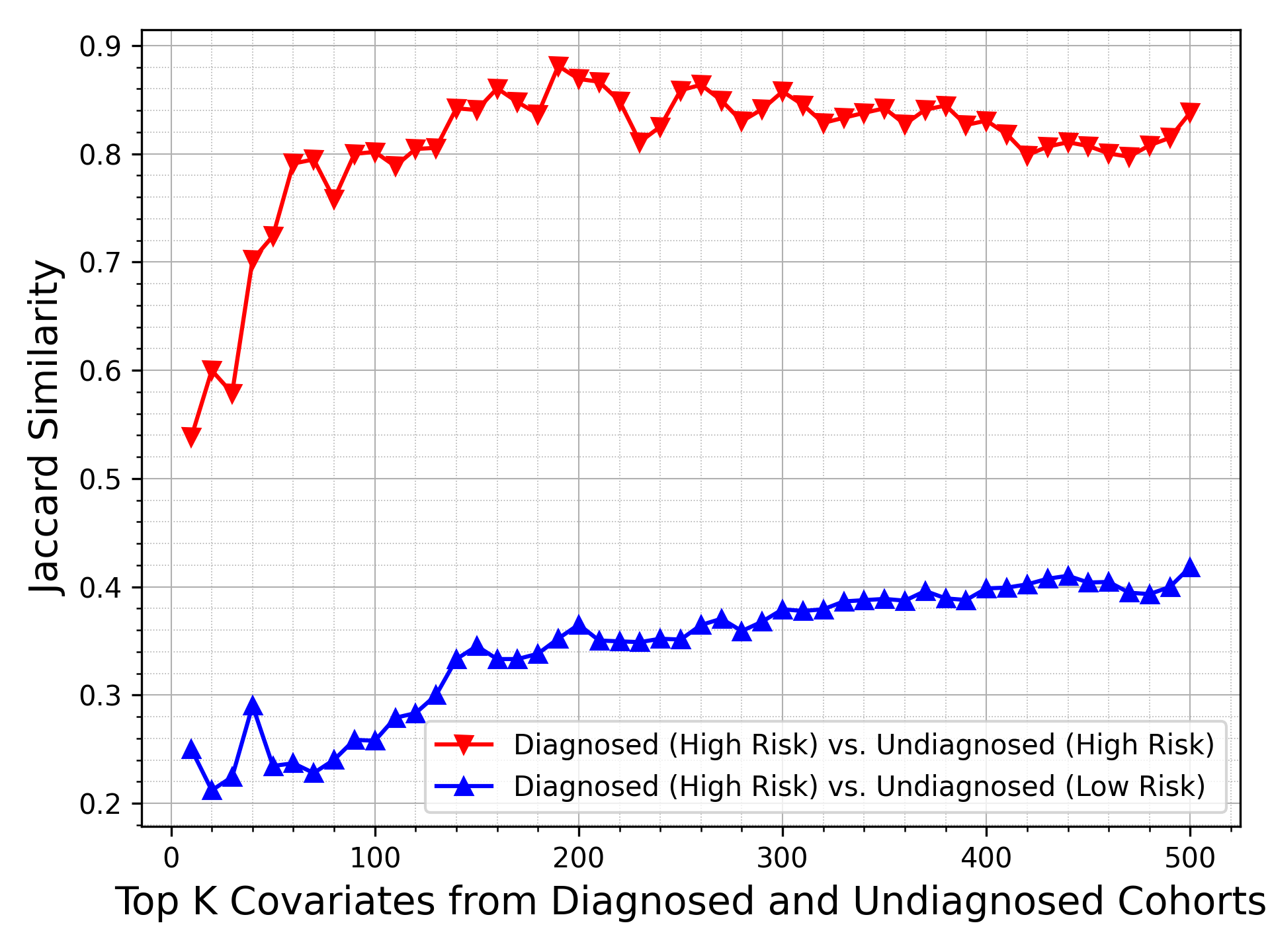}
\caption{\textbf{Jaccard similarity of top 10-500 most frequent covariates in diagnosed vs. undiagnosed.} \textcolor{red}{Red line:} high-risk diagnosed vs. high-risk undiagnosed. \textcolor{blue}{Blue line:} high-risk diagnosed vs. low-risk undiagnosed. High-risk: model estimated risk score $\geq$0.90, low-risk: model estimated risk score $\leq$0.10.}
\label{fig:jacc_sim}
\end{figure}

As anticipated (Figure \ref{fig:jacc_sim}), diagnosed and undiagnosed individuals with high predicted risk of T2DM have substantial overlap in their covariate patterns (Jaccard similarity: 0.54-0.88). This supports our hypothesis that these groups share similar clinical profiles. In contrast, the covariate overlap between high-risk diagnosed individuals and low-risk undiagnosed individuals is considerably lower (Jaccard similarity: 0.21–0.42), indicating distinct clinical characteristics. These findings demonstrate the model’s ability to identify undiagnosed individuals at elevated risk by leveraging covariate patterns learned from the diagnosed population.

\subsection{Validation Method 4}
As expected, model performance (Table \ref{tab:validation_method_4}) declined as the risk score threshold for selecting negative examples increased. The most significant drop occurred when negative examples were randomly selected from the entire undiagnosed population (threshold $\leq$1.0). This decline is likely due to the inclusion of high-risk undiagnosed individuals who may in fact be undiagnosed T2DM cases (i.e., false negatives). These findings support the hypothesis that undiagnosed individuals assigned low risk scores by our model are more likely to be true negatives, allowing the classifier to achieve its highest performance when trained on this cleaner subset (threshold $\leq$0.1).

\begin{table*}[h]
\caption{
\textbf{Classification performance of XGBoost models trained on balanced datasets.} 
\textnormal{Reported performance metrics represent the mean and 95\% CI across 40 iterations. MCC: Matthews Correlation Coefficient, AUC-ROC: Area Under the Curve of the Receiver Operating Characteristic, BS: Brier score loss, F1: F-measure.}}
\makebox[\textwidth][c]{%
\resizebox{\textwidth}{!}{%
\begin{tabular}{|c|c|c|c|c|c|}
\hline
\textbf{\begin{tabular}[c]{@{}c@{}}Risk\\ Threshold\end{tabular}} & \textbf{Accuracy} & \textbf{MCC} & \textbf{AUC-ROC} & \textbf{BS} & \textbf{F1} \\ \hline
0.1 & \begin{tabular}[c]{@{}c@{}}0.9091\\ {[}0.9085, 0.9096{]}\end{tabular} & \begin{tabular}[c]{@{}c@{}}0.8262\\ {[}0.8251, 0.8272{]}\end{tabular} & \begin{tabular}[c]{@{}c@{}}0.9702\\ {[}0.9698, 0.9707{]}\end{tabular} & \begin{tabular}[c]{@{}c@{}}0.0655\\ {[}0.0651, 0.0659{]}\end{tabular} & \begin{tabular}[c]{@{}c@{}}0.9022\\ {[}0.9016, 0.9028{]}\end{tabular} \\ \hline
0.5 & \begin{tabular}[c]{@{}c@{}}0.8831\\ {[}0.8826, 0.8836{]}\end{tabular} & \begin{tabular}[c]{@{}c@{}}0.7684\\ {[}0.7674, 0.7693{]}\end{tabular} & \begin{tabular}[c]{@{}c@{}}0.9547\\ {[}0.9544, 0.9550{]}\end{tabular} & \begin{tabular}[c]{@{}c@{}}0.0847\\ {[}0.0845, 0.085{]}\end{tabular} & \begin{tabular}[c]{@{}c@{}}0.8786\\ {[}0.8781, 0.8791{]}\end{tabular} \\ \hline
1.0 & \begin{tabular}[c]{@{}c@{}}0.8207\\ {[}0.8200, 0.8212{]}\end{tabular} & \begin{tabular}[c]{@{}c@{}}0.6423\\ {[}0.6410, 0.6432{]}\end{tabular} & \begin{tabular}[c]{@{}c@{}}0.9024\\ {[}0.9020, 0.9028{]}\end{tabular} & \begin{tabular}[c]{@{}c@{}}0.1274\\ {[}0.1271, 0.1277{]}\end{tabular} & \begin{tabular}[c]{@{}c@{}}0.8252\\ {[}0.8246, 0.8258{]}\end{tabular} \\ \hline
\end{tabular}
}
}
\label{tab:validation_method_4}
\end{table*}

\section{Discussion and Conclusions}

T2DM is associated with several complications, including diabetic ketoacidosis, diabetic coma, hypoglycemia, hyperglycemia, diabetic foot, diabetic neuropathy, cardiovascular disease, kidney damage, vision loss, congestive heart failure, and arterial hypertension \citep{jbhi2025_1}. Early identification of T2DM is crucial, as it enables timely intervention to slow disease progression, reduce the incidence and severity of complications, lower healthcare costs, and support both individualized patient care and broader public health initiatives. Ultimately, early detection and management contribute to improved patient outcomes and quality of life.

This study presents a novel approach that leverages clinical patterns from diagnosed patients to estimate T2DM risk in undiagnosed individuals. Among the undiagnosed group, 3.18\% overall and 4.67\% among adults aged $\geq$18 years were classified as high risk. These estimates align with CDC-reported prevalence of undiagnosed diabetes among adults aged $\geq$18 years in 2017–2020 [3.4\% (2.7–4.2\%)] \citep{jbhi2025_6}. These findings highlight the method’s potential to identify high-risk individuals who could benefit from early screening and intervention.

In real-world clinical and epidemiological settings, medical records are rarely comprehensively annotated; labeling efforts typically focus only on confirmed cases \citep{jbhi2025_40, jbhi2025_41, jbhi2025_42}. This poses challenges for conventional binary classifiers, which require labeled examples of both positive and negative cases. Furthermore, the limited number of labeled cases often leads to class imbalance, further affecting model performance \citep{jbhi2025_43}. Our approach addresses these limitations by eliminating the need for annotated negatives, enabling risk stratification using only confirmed positives. Although demonstrated with T2DM, the method is generalizable to other phenotypes. Validation through multiple experiments and expert review shows strong agreement between model-derived risk scores and clinical judgment. As demonstrated by the performance of the XGBoost classifier (Table \ref{tab:validation_method_4}), the method can also reliably identify negative cases using a low-risk threshold.

Our Rank-Weighted Coefficient (RWC) approach offers an innovative method for identifying important covariates by aggregating their contributions across individual components of NMF. In our study, covariates with high RWC values—such as end-stage renal disease (ESRD), hypertension, obstructive sleep apnea (OSA), hyperlipidemia, obesity, and the use of hydrochlorothiazide (HCTZ)—show strong associations with T2DM, consistent with prior research. For example, OSA is linked to metabolic dysregulation \citep{jbhi2025_44}; T2DM is a leading cause of ESRD \citep{jbhi2025_45}; hypertension frequently co-occurs with T2DM due to shared pathophysiology \citep{jbhi2025_46}; hyperlipidemia and obesity contribute to insulin resistance and increased T2DM risk \citep{jbhi2025_47}; and although HCTZ and other related thiazide diuretics are widely used for hypertension management, they have the potential to worsen metabolic abnormalities in existing T2D patients.  Additionally, treatment with thiazide diuretics is known to contribute to the development of diabetes, especially if these medications are taken without careful lab monitoring and medical oversight \citep{jbhi2025_48}. Notably, both the AI and SME also recognized these covariates as important in their classifications. However, the AI appeared to assign greater weight to these features, whereas SME incorporated broader clinical context, leading to some divergence in classification outcomes (Table \ref{tab:validation_method_2}).

Overall, as demonstrated by our results, the proposed unsupervised framework is scalable and label-efficient for identifying T2DM risk, with strong potential for application to other phenotypes in real-world settings where underdiagnosis and limited labeling are prevalent.

\subsection{Limitations and Future Work}
Our study is limited by its reliance on a single dataset, which may not be representative of all populations or clinical scenarios. Future work will focus on validating our findings across multiple datasets to assess the generalizability and robustness of our approach. We selected the number of NMF components based on reconstruction error and the number of broader ICD-10-CM categories related to T2DM. While reconstruction error is a commonly used heuristic, it may not always yield the most optimal component count. In our experiments with 5 to 9 components, the list of important features remained stable despite variations in RWC values. Due to the lack of hemoglobin A1C (HbA1C) measurements in claims data, we relied on SME input to set percentile-based risk thresholds. Future integration of EHR data containing HbA1C values will allow for more precise thresholding and potentially improve risk estimation accuracy.

\bmhead{Acknowledgements} This research was supported by funding from the US National Library of Medicine grant R00-LM13367. The views expressed in this paper are those of the authors and do not necessarily reflect those of the National Institutes of Health.

\begin{appendices}

\end{appendices}

\bibliographystyle{bst/sn-mathphys-num}
\bibliography{sn-bibliography}


\begin{thebibliography}{43}
\ifx \bisbn   \undefined \def \bisbn  #1{ISBN #1}\fi
\ifx \binits  \undefined \def \binits#1{#1}\fi
\ifx \bauthor  \undefined \def \bauthor#1{#1}\fi
\ifx \batitle  \undefined \def \batitle#1{#1}\fi
\ifx \bjtitle  \undefined \def \bjtitle#1{#1}\fi
\ifx \bvolume  \undefined \def \bvolume#1{\textbf{#1}}\fi
\ifx \byear  \undefined \def \byear#1{#1}\fi
\ifx \bissue  \undefined \def \bissue#1{#1}\fi
\ifx \bfpage  \undefined \def \bfpage#1{#1}\fi
\ifx \blpage  \undefined \def \blpage #1{#1}\fi
\ifx \burl  \undefined \def \burl#1{\textsf{#1}}\fi
\ifx \doiurl  \undefined \def \doiurl#1{\url{https://doi.org/#1}}\fi
\ifx \betal  \undefined \def \betal{\textit{et al.}}\fi
\ifx \binstitute  \undefined \def \binstitute#1{#1}\fi
\ifx \binstitutionaled  \undefined \def \binstitutionaled#1{#1}\fi
\ifx \bctitle  \undefined \def \bctitle#1{#1}\fi
\ifx \beditor  \undefined \def \beditor#1{#1}\fi
\ifx \bpublisher  \undefined \def \bpublisher#1{#1}\fi
\ifx \bbtitle  \undefined \def \bbtitle#1{#1}\fi
\ifx \bedition  \undefined \def \bedition#1{#1}\fi
\ifx \bseriesno  \undefined \def \bseriesno#1{#1}\fi
\ifx \blocation  \undefined \def \blocation#1{#1}\fi
\ifx \bsertitle  \undefined \def \bsertitle#1{#1}\fi
\ifx \bsnm \undefined \def \bsnm#1{#1}\fi
\ifx \bsuffix \undefined \def \bsuffix#1{#1}\fi
\ifx \bparticle \undefined \def \bparticle#1{#1}\fi
\ifx \barticle \undefined \def \barticle#1{#1}\fi
\bibcommenthead
\ifx \bconfdate \undefined \def \bconfdate #1{#1}\fi
\ifx \botherref \undefined \def \botherref #1{#1}\fi
\ifx \url \undefined \def \url#1{\textsf{#1}}\fi
\ifx \bchapter \undefined \def \bchapter#1{#1}\fi
\ifx \bbook \undefined \def \bbook#1{#1}\fi
\ifx \bcomment \undefined \def \bcomment#1{#1}\fi
\ifx \oauthor \undefined \def \oauthor#1{#1}\fi
\ifx \citeauthoryear \undefined \def \citeauthoryear#1{#1}\fi
\ifx \endbibitem  \undefined \def \endbibitem {}\fi
\ifx \bconflocation  \undefined \def \bconflocation#1{#1}\fi
\ifx \arxivurl  \undefined \def \arxivurl#1{\textsf{#1}}\fi
\csname PreBibitemsHook\endcsname

\bibitem[\protect\citeauthoryear{Farmaki et~al.}{2020}]{jbhi2025_1}
\begin{barticle}
\bauthor{\bsnm{Farmaki}, \binits{P.}},
\bauthor{\bsnm{Damaskos}, \binits{C.}},
\bauthor{\bsnm{Garmpis}, \binits{N.}},
\bauthor{\bsnm{Garmpi}, \binits{A.}},
\bauthor{\bsnm{Savvanis}, \binits{S.}},
\bauthor{\bsnm{Diamantis}, \binits{E.}}:
\batitle{Complications of the type 2 diabetes mellitus}.
\bjtitle{Current cardiology reviews}
\bvolume{16}(\bissue{4}),
\bfpage{249}--\blpage{251}
(\byear{2020})
\end{barticle}
\endbibitem

\bibitem[\protect\citeauthoryear{(WHO)}{}]{jbhi2025_2}
\begin{botherref}
\oauthor{\bsnm{(WHO)}, \binits{W.H.O.}}:
Diabetes overview.
Available at \url{https://www.who.int/news-room/fact-sheets/detail/diabetes}.
Accessed April 11, 2025
\end{botherref}
\endbibitem

\bibitem[\protect\citeauthoryear{(IDF)}{}]{jbhi2025_4}
\begin{botherref}
\oauthor{\bsnm{(IDF)}, \binits{T.I.D.F.}}:
Diabetes facts and figures.
Available at \url{https://idf.org/about-diabetes/diabetes-facts-figures/}.
Accessed April 11, 2025
\end{botherref}
\endbibitem

\bibitem[\protect\citeauthoryear{for Disease~Control and (CDC)}{}]{jbhi2025_5}
\begin{botherref}
\oauthor{\bsnm{Disease~Control}, \binits{C.}},
\oauthor{\bsnm{(CDC)}, \binits{P.}}:
About Underlying Cause of Death, 2018-2022.
Available at \url{https://wonder.cdc.gov/controller/saved/D158/D389F360}.
Accessed April 11, 2025
\end{botherref}
\endbibitem

\bibitem[\protect\citeauthoryear{for Disease~Control and Prevention}{}]{jbhi2025_6}
\begin{botherref}
\oauthor{\bsnm{Disease~Control}, \binits{C.}},
\oauthor{\bsnm{Prevention}}:
National Diabetes Statistics Report.
Available at \url{https://www.cdc.gov/diabetes/php/data-research/index.html}.
Accessed April 11, 2025
\end{botherref}
\endbibitem

\bibitem[\protect\citeauthoryear{Parker et~al.}{2024}]{jbhi2025_7}
\begin{barticle}
\bauthor{\bsnm{Parker}, \binits{E.D.}},
\bauthor{\bsnm{Lin}, \binits{J.}},
\bauthor{\bsnm{Mahoney}, \binits{T.}},
\bauthor{\bsnm{Ume}, \binits{N.}},
\bauthor{\bsnm{Yang}, \binits{G.}},
\bauthor{\bsnm{Gabbay}, \binits{R.A.}},
\bauthor{\bsnm{ElSayed}, \binits{N.A.}},
\bauthor{\bsnm{Bannuru}, \binits{R.R.}}:
\batitle{Economic costs of diabetes in the us in 2022}.
\bjtitle{Diabetes care}
\bvolume{47}(\bissue{1}),
\bfpage{26}--\blpage{43}
(\byear{2024})
\end{barticle}
\endbibitem

\bibitem[\protect\citeauthoryear{Zou et~al.}{2018}]{jbhi2025_8}
\begin{barticle}
\bauthor{\bsnm{Zou}, \binits{Q.}},
\bauthor{\bsnm{Qu}, \binits{K.}},
\bauthor{\bsnm{Luo}, \binits{Y.}},
\bauthor{\bsnm{Yin}, \binits{D.}},
\bauthor{\bsnm{Ju}, \binits{Y.}},
\bauthor{\bsnm{Tang}, \binits{H.}}:
\batitle{Predicting diabetes mellitus with machine learning techniques}.
\bjtitle{Frontiers in genetics}
\bvolume{9},
\bfpage{515}
(\byear{2018})
\end{barticle}
\endbibitem

\bibitem[\protect\citeauthoryear{Breiman}{2001}]{jbhi2025_12}
\begin{barticle}
\bauthor{\bsnm{Breiman}, \binits{L.}}:
\batitle{Random forests}.
\bjtitle{Machine learning}
\bvolume{45},
\bfpage{5}--\blpage{32}
(\byear{2001})
\end{barticle}
\endbibitem

\bibitem[\protect\citeauthoryear{Rumelhart et~al.}{1986}]{jbhi2025_26}
\begin{barticle}
\bauthor{\bsnm{Rumelhart}, \binits{D.E.}},
\bauthor{\bsnm{Hinton}, \binits{G.E.}},
\bauthor{\bsnm{Williams}, \binits{R.J.}}:
\batitle{Learning representations by back-propagating errors}.
\bjtitle{nature}
\bvolume{323}(\bissue{6088}),
\bfpage{533}--\blpage{536}
(\byear{1986})
\end{barticle}
\endbibitem

\bibitem[\protect\citeauthoryear{Quinlan}{1986}]{jbhi2025_25}
\begin{barticle}
\bauthor{\bsnm{Quinlan}, \binits{J.R.}}:
\batitle{Induction of decision trees}.
\bjtitle{Machine learning}
\bvolume{1},
\bfpage{81}--\blpage{106}
(\byear{1986})
\end{barticle}
\endbibitem

\bibitem[\protect\citeauthoryear{Lai et~al.}{2019}]{jbhi2025_9}
\begin{barticle}
\bauthor{\bsnm{Lai}, \binits{H.}},
\bauthor{\bsnm{Huang}, \binits{H.}},
\bauthor{\bsnm{Keshavjee}, \binits{K.}},
\bauthor{\bsnm{Guergachi}, \binits{A.}},
\bauthor{\bsnm{Gao}, \binits{X.}}:
\batitle{Predictive models for diabetes mellitus using machine learning techniques}.
\bjtitle{BMC endocrine disorders}
\bvolume{19},
\bfpage{1}--\blpage{9}
(\byear{2019})
\end{barticle}
\endbibitem

\bibitem[\protect\citeauthoryear{Friedman}{2001}]{jbhi2025_27}
\begin{botherref}
\oauthor{\bsnm{Friedman}, \binits{J.H.}}:
Greedy function approximation: a gradient boosting machine.
Annals of statistics,
1189--1232
(2001)
\end{botherref}
\endbibitem

\bibitem[\protect\citeauthoryear{Cox}{1958}]{jbhi2025_28}
\begin{barticle}
\bauthor{\bsnm{Cox}, \binits{D.R.}}:
\batitle{The regression analysis of binary sequences}.
\bjtitle{Journal of the Royal Statistical Society Series B: Statistical Methodology}
\bvolume{20}(\bissue{2}),
\bfpage{215}--\blpage{232}
(\byear{1958})
\end{barticle}
\endbibitem

\bibitem[\protect\citeauthoryear{Kopitar et~al.}{2020}]{jbhi2025_10}
\begin{barticle}
\bauthor{\bsnm{Kopitar}, \binits{L.}},
\bauthor{\bsnm{Kocbek}, \binits{P.}},
\bauthor{\bsnm{Cilar}, \binits{L.}},
\bauthor{\bsnm{Sheikh}, \binits{A.}},
\bauthor{\bsnm{Stiglic}, \binits{G.}}:
\batitle{Early detection of type 2 diabetes mellitus using machine learning-based prediction models}.
\bjtitle{Scientific reports}
\bvolume{10}(\bissue{1}),
\bfpage{11981}
(\byear{2020})
\end{barticle}
\endbibitem

\bibitem[\protect\citeauthoryear{Friedman et~al.}{2010}]{jbhi2025_11}
\begin{barticle}
\bauthor{\bsnm{Friedman}, \binits{J.H.}},
\bauthor{\bsnm{Hastie}, \binits{T.}},
\bauthor{\bsnm{Tibshirani}, \binits{R.}}:
\batitle{Regularization paths for generalized linear models via coordinate descent}.
\bjtitle{Journal of statistical software}
\bvolume{33},
\bfpage{1}--\blpage{22}
(\byear{2010})
\end{barticle}
\endbibitem

\bibitem[\protect\citeauthoryear{Chen and Guestrin}{2016}]{jbhi2025_13}
\begin{bchapter}
\bauthor{\bsnm{Chen}, \binits{T.}},
\bauthor{\bsnm{Guestrin}, \binits{C.}}:
\bctitle{Xgboost: A scalable tree boosting system}.
In: \bbtitle{Proceedings of the 22nd Acm Sigkdd International Conference on Knowledge Discovery and Data Mining},
pp. \bfpage{785}--\blpage{794}
(\byear{2016})
\end{bchapter}
\endbibitem

\bibitem[\protect\citeauthoryear{Ke et~al.}{2017}]{jbhi2025_14}
\begin{botherref}
\oauthor{\bsnm{Ke}, \binits{G.}},
\oauthor{\bsnm{Meng}, \binits{Q.}},
\oauthor{\bsnm{Finley}, \binits{T.}},
\oauthor{\bsnm{Wang}, \binits{T.}},
\oauthor{\bsnm{Chen}, \binits{W.}},
\oauthor{\bsnm{Ma}, \binits{W.}},
\oauthor{\bsnm{Ye}, \binits{Q.}},
\oauthor{\bsnm{Liu}, \binits{T.-Y.}}:
Lightgbm: A highly efficient gradient boosting decision tree.
Advances in neural information processing systems
\textbf{30}
(2017)
\end{botherref}
\endbibitem

\bibitem[\protect\citeauthoryear{Cover and Hart}{1967}]{jbhi2025_16}
\begin{barticle}
\bauthor{\bsnm{Cover}, \binits{T.}},
\bauthor{\bsnm{Hart}, \binits{P.}}:
\batitle{Nearest neighbor pattern classification}.
\bjtitle{IEEE transactions on information theory}
\bvolume{13}(\bissue{1}),
\bfpage{21}--\blpage{27}
(\byear{1967})
\end{barticle}
\endbibitem

\bibitem[\protect\citeauthoryear{Breiman et~al.}{2017}]{jbhi2025_17}
\begin{bbook}
\bauthor{\bsnm{Breiman}, \binits{L.}},
\bauthor{\bsnm{Friedman}, \binits{J.}},
\bauthor{\bsnm{Olshen}, \binits{R.A.}},
\bauthor{\bsnm{Stone}, \binits{C.J.}}:
\bbtitle{Classification and Regression Trees}.
\bpublisher{Routledge}, \blocation{???}
(\byear{2017})
\end{bbook}
\endbibitem

\bibitem[\protect\citeauthoryear{Cortes and Vapnik}{1995}]{jbhi2025_18}
\begin{barticle}
\bauthor{\bsnm{Cortes}, \binits{C.}},
\bauthor{\bsnm{Vapnik}, \binits{V.}}:
\batitle{Support-vector networks}.
\bjtitle{Machine learning}
\bvolume{20},
\bfpage{273}--\blpage{297}
(\byear{1995})
\end{barticle}
\endbibitem

\bibitem[\protect\citeauthoryear{Mani et~al.}{2012}]{jbhi2025_15}
\begin{bchapter}
\bauthor{\bsnm{Mani}, \binits{S.}},
\bauthor{\bsnm{Chen}, \binits{Y.}},
\bauthor{\bsnm{Elasy}, \binits{T.}},
\bauthor{\bsnm{Clayton}, \binits{W.}},
\bauthor{\bsnm{Denny}, \binits{J.}}:
\bctitle{Type 2 diabetes risk forecasting from emr data using machine learning}.
In: \bbtitle{AMIA Annual Symposium Proceedings},
vol. \bseriesno{2012},
p. \bfpage{606}
(\byear{2012})
\end{bchapter}
\endbibitem

\bibitem[\protect\citeauthoryear{Razavian et~al.}{2015}]{jbhi2025_20}
\begin{barticle}
\bauthor{\bsnm{Razavian}, \binits{N.}},
\bauthor{\bsnm{Blecker}, \binits{S.}},
\bauthor{\bsnm{Schmidt}, \binits{A.M.}},
\bauthor{\bsnm{Smith-McLallen}, \binits{A.}},
\bauthor{\bsnm{Nigam}, \binits{S.}},
\bauthor{\bsnm{Sontag}, \binits{D.}}:
\batitle{Population-level prediction of type 2 diabetes from claims data and analysis of risk factors}.
\bjtitle{Big Data}
\bvolume{3}(\bissue{4}),
\bfpage{277}--\blpage{287}
(\byear{2015})
\end{barticle}
\endbibitem

\bibitem[\protect\citeauthoryear{Ravaut et~al.}{2021}]{jbhi2025_21}
\begin{barticle}
\bauthor{\bsnm{Ravaut}, \binits{M.}},
\bauthor{\bsnm{Harish}, \binits{V.}},
\bauthor{\bsnm{Sadeghi}, \binits{H.}},
\bauthor{\bsnm{Leung}, \binits{K.K.}},
\bauthor{\bsnm{Volkovs}, \binits{M.}},
\bauthor{\bsnm{Kornas}, \binits{K.}},
\bauthor{\bsnm{Watson}, \binits{T.}},
\bauthor{\bsnm{Poutanen}, \binits{T.}},
\bauthor{\bsnm{Rosella}, \binits{L.C.}}:
\batitle{Development and validation of a machine learning model using administrative health data to predict onset of type 2 diabetes}.
\bjtitle{JAMA network open}
\bvolume{4}(\bissue{5}),
\bfpage{2111315}--\blpage{2111315}
(\byear{2021})
\end{barticle}
\endbibitem

\bibitem[\protect\citeauthoryear{Klein~Woolthuis et~al.}{2007}]{jbhi2025_19}
\begin{barticle}
\bauthor{\bsnm{Klein~Woolthuis}, \binits{E.P.}},
\bauthor{\bsnm{Grauw}, \binits{W.J.}},
\bauthor{\bsnm{Gerwen}, \binits{W.H.}},
\bauthor{\bsnm{Hoogen}, \binits{H.J.}},
\bauthor{\bsnm{Lisdonk}, \binits{E.H.}},
\bauthor{\bsnm{Metsemakers}, \binits{J.F.}},
\bauthor{\bsnm{Weel}, \binits{C.}}:
\batitle{Identifying people at risk for undiagnosed type 2 diabetes using the gp's electronic medical record}.
\bjtitle{Family practice}
\bvolume{24}(\bissue{3}),
\bfpage{230}--\blpage{236}
(\byear{2007})
\end{barticle}
\endbibitem

\bibitem[\protect\citeauthoryear{McInnes et~al.}{2018}]{jbhi2025_23}
\begin{botherref}
\oauthor{\bsnm{McInnes}, \binits{L.}},
\oauthor{\bsnm{Healy}, \binits{J.}},
\oauthor{\bsnm{Melville}, \binits{J.}}:
Umap: Uniform manifold approximation and projection for dimension reduction.
arXiv preprint arXiv:1802.03426
(2018)
\end{botherref}
\endbibitem

\bibitem[\protect\citeauthoryear{Ester et~al.}{1996}]{jbhi2025_24}
\begin{bchapter}
\bauthor{\bsnm{Ester}, \binits{M.}},
\bauthor{\bsnm{Kriegel}, \binits{H.-P.}},
\bauthor{\bsnm{Sander}, \binits{J.}},
\bauthor{\bsnm{Xu}, \binits{X.}}, \betal:
\bctitle{A density-based algorithm for discovering clusters in large spatial databases with noise}.
In: \bbtitle{Kdd},
vol. \bseriesno{96},
pp. \bfpage{226}--\blpage{231}
(\byear{1996})
\end{bchapter}
\endbibitem

\bibitem[\protect\citeauthoryear{Bej et~al.}{2022}]{jbhi2025_22}
\begin{barticle}
\bauthor{\bsnm{Bej}, \binits{S.}},
\bauthor{\bsnm{Sarkar}, \binits{J.}},
\bauthor{\bsnm{Biswas}, \binits{S.}},
\bauthor{\bsnm{Mitra}, \binits{P.}},
\bauthor{\bsnm{Chakrabarti}, \binits{P.}},
\bauthor{\bsnm{Wolkenhauer}, \binits{O.}}:
\batitle{Identification and epidemiological characterization of type-2 diabetes sub-population using an unsupervised machine learning approach}.
\bjtitle{Nutrition \& Diabetes}
\bvolume{12}(\bissue{1}),
\bfpage{27}
(\byear{2022})
\end{barticle}
\endbibitem

\bibitem[\protect\citeauthoryear{Voss et~al.}{2015}]{jbhi2025_29}
\begin{barticle}
\bauthor{\bsnm{Voss}, \binits{E.A.}},
\bauthor{\bsnm{Makadia}, \binits{R.}},
\bauthor{\bsnm{Matcho}, \binits{A.}},
\bauthor{\bsnm{Ma}, \binits{Q.}},
\bauthor{\bsnm{Knoll}, \binits{C.}},
\bauthor{\bsnm{Schuemie}, \binits{M.}},
\bauthor{\bsnm{DeFalco}, \binits{F.J.}},
\bauthor{\bsnm{Londhe}, \binits{A.}},
\bauthor{\bsnm{Zhu}, \binits{V.}},
\bauthor{\bsnm{Ryan}, \binits{P.B.}}:
\batitle{Feasibility and utility of applications of the common data model to multiple, disparate observational health databases}.
\bjtitle{Journal of the American Medical Informatics Association}
\bvolume{22}(\bissue{3}),
\bfpage{553}--\blpage{564}
(\byear{2015})
\end{barticle}
\endbibitem

\bibitem[\protect\citeauthoryear{Fr{\'e}nay and Verleysen}{2013}]{jbhi2025_33}
\begin{barticle}
\bauthor{\bsnm{Fr{\'e}nay}, \binits{B.}},
\bauthor{\bsnm{Verleysen}, \binits{M.}}:
\batitle{Classification in the presence of label noise: a survey}.
\bjtitle{IEEE transactions on neural networks and learning systems}
\bvolume{25}(\bissue{5}),
\bfpage{845}--\blpage{869}
(\byear{2013})
\end{barticle}
\endbibitem

\bibitem[\protect\citeauthoryear{Kumar and Lambert}{2024}]{jbhi2025_31}
\begin{barticle}
\bauthor{\bsnm{Kumar}, \binits{P.}},
\bauthor{\bsnm{Lambert}, \binits{C.G.}}:
\batitle{Positive unlabeled learning selected not at random (pulsnar): class proportion estimation without the selected completely at random assumption}.
\bjtitle{PeerJ Computer Science}
\bvolume{10},
\bfpage{2451}
(\byear{2024})
\end{barticle}
\endbibitem

\bibitem[\protect\citeauthoryear{Jaskie et~al.}{2019}]{jbhi2025_32}
\begin{bchapter}
\bauthor{\bsnm{Jaskie}, \binits{K.}},
\bauthor{\bsnm{Elkan}, \binits{C.}},
\bauthor{\bsnm{Spanias}, \binits{A.}}:
\bctitle{A modified logistic regression for positive and unlabeled learning}.
In: \bbtitle{2019 53rd Asilomar Conference on Signals, Systems, and Computers},
pp. \bfpage{2007}--\blpage{2011}
(\byear{2019}).
\bcomment{IEEE}
\end{bchapter}
\endbibitem

\bibitem[\protect\citeauthoryear{Lee and Seung}{1999}]{jbhi2025_34}
\begin{barticle}
\bauthor{\bsnm{Lee}, \binits{D.D.}},
\bauthor{\bsnm{Seung}, \binits{H.S.}}:
\batitle{Learning the parts of objects by non-negative matrix factorization}.
\bjtitle{nature}
\bvolume{401}(\bissue{6755}),
\bfpage{788}--\blpage{791}
(\byear{1999})
\end{barticle}
\endbibitem

\bibitem[\protect\citeauthoryear{Kullback and Leibler}{1951}]{jbhi2025_35}
\begin{barticle}
\bauthor{\bsnm{Kullback}, \binits{S.}},
\bauthor{\bsnm{Leibler}, \binits{R.A.}}:
\batitle{On information and sufficiency}.
\bjtitle{The annals of mathematical statistics}
\bvolume{22}(\bissue{1}),
\bfpage{79}--\blpage{86}
(\byear{1951})
\end{barticle}
\endbibitem

\bibitem[\protect\citeauthoryear{ICDData}{}]{jbhi2025_30}
\begin{botherref}
\oauthor{\bsnm{ICDData}}:
Type 2 diabetes mellitus E11.
Available at \url{https://www.icd10data.com/ICD10CM/Codes/E00-E89/E08-E13/E11-}.
Accessed April 11, 2025
\end{botherref}
\endbibitem

\bibitem[\protect\citeauthoryear{Kumar et~al.}{2024}]{jbhi2025_40}
\begin{botherref}
\oauthor{\bsnm{Kumar}, \binits{P.}},
\oauthor{\bsnm{Moomtaheen}, \binits{F.}},
\oauthor{\bsnm{Malec}, \binits{S.A.}},
\oauthor{\bsnm{Yang}, \binits{J.J.}},
\oauthor{\bsnm{Bologa}, \binits{C.G.}},
\oauthor{\bsnm{Schneider}, \binits{K.A.}},
\oauthor{\bsnm{Zhu}, \binits{Y.}},
\oauthor{\bsnm{Tohen}, \binits{M.}},
\oauthor{\bsnm{Villarreal}, \binits{G.}},
\oauthor{\bsnm{Perkins}, \binits{D.J.}}, et al.:
Detecting opioid use disorder in health claims data with positive unlabeled learning.
IEEE Journal of Biomedical and Health Informatics
(2024)
\end{botherref}
\endbibitem

\bibitem[\protect\citeauthoryear{Kumar et~al.}{2020}]{jbhi2025_41}
\begin{barticle}
\bauthor{\bsnm{Kumar}, \binits{P.}},
\bauthor{\bsnm{Nestsiarovich}, \binits{A.}},
\bauthor{\bsnm{Nelson}, \binits{S.J.}},
\bauthor{\bsnm{Kerner}, \binits{B.}},
\bauthor{\bsnm{Perkins}, \binits{D.J.}},
\bauthor{\bsnm{Lambert}, \binits{C.G.}}:
\batitle{Imputation and characterization of uncoded self-harm in major mental illness using machine learning}.
\bjtitle{Journal of the American Medical Informatics Association}
\bvolume{27}(\bissue{1}),
\bfpage{136}--\blpage{146}
(\byear{2020})
\end{barticle}
\endbibitem

\bibitem[\protect\citeauthoryear{Lin et~al.}{2015}]{jbhi2025_42}
\begin{barticle}
\bauthor{\bsnm{Lin}, \binits{P.-J.}},
\bauthor{\bsnm{Kent}, \binits{D.M.}},
\bauthor{\bsnm{Winn}, \binits{A.}},
\bauthor{\bsnm{Cohen}, \binits{J.T.}},
\bauthor{\bsnm{Neumann}, \binits{P.J.}}, \betal:
\batitle{Multiple chronic conditions in type 2 diabetes mellitus: prevalence and consequences}.
\bjtitle{Am J Manag Care}
\bvolume{21}(\bissue{1}),
\bfpage{23}--\blpage{34}
(\byear{2015})
\end{barticle}
\endbibitem

\bibitem[\protect\citeauthoryear{Luque et~al.}{2019}]{jbhi2025_43}
\begin{barticle}
\bauthor{\bsnm{Luque}, \binits{A.}},
\bauthor{\bsnm{Carrasco}, \binits{A.}},
\bauthor{\bsnm{Mart{\'\i}n}, \binits{A.}},
\bauthor{\bsnm{Las~Heras}, \binits{A.}}:
\batitle{The impact of class imbalance in classification performance metrics based on the binary confusion matrix}.
\bjtitle{Pattern Recognition}
\bvolume{91},
\bfpage{216}--\blpage{231}
(\byear{2019})
\end{barticle}
\endbibitem

\bibitem[\protect\citeauthoryear{Choi et~al.}{2019}]{jbhi2025_44}
\begin{barticle}
\bauthor{\bsnm{Choi}, \binits{H.S.}},
\bauthor{\bsnm{Kim}, \binits{H.Y.}},
\bauthor{\bsnm{Han}, \binits{K.-D.}},
\bauthor{\bsnm{Jung}, \binits{J.-H.}},
\bauthor{\bsnm{Kim}, \binits{C.S.}},
\bauthor{\bsnm{Bae}, \binits{E.H.}},
\bauthor{\bsnm{Ma}, \binits{S.K.}},
\bauthor{\bsnm{Kim}, \binits{S.W.}}:
\batitle{Obstructive sleep apnea as a risk factor for incident end stage renal disease: a nationwide population-based cohort study from korea}.
\bjtitle{Clinical and experimental nephrology}
\bvolume{23},
\bfpage{1391}--\blpage{1397}
(\byear{2019})
\end{barticle}
\endbibitem

\bibitem[\protect\citeauthoryear{Keane et~al.}{2003}]{jbhi2025_45}
\begin{barticle}
\bauthor{\bsnm{Keane}, \binits{W.F.}},
\bauthor{\bsnm{Brenner}, \binits{B.M.}},
\bauthor{\bsnm{De~Zeeuw}, \binits{D.}},
\bauthor{\bsnm{Grunfeld}, \binits{J.-P.}},
\bauthor{\bsnm{McGill}, \binits{J.}},
\bauthor{\bsnm{Mitch}, \binits{W.E.}},
\bauthor{\bsnm{Ribeiro}, \binits{A.B.}},
\bauthor{\bsnm{Shahinfar}, \binits{S.}},
\bauthor{\bsnm{Simpson}, \binits{R.L.}},
\bauthor{\bsnm{Snapinn}, \binits{S.M.}}, \betal:
\batitle{The risk of developing end-stage renal disease in patients with type 2 diabetes and nephropathy: the renaal study}.
\bjtitle{Kidney international}
\bvolume{63}(\bissue{4}),
\bfpage{1499}--\blpage{1507}
(\byear{2003})
\end{barticle}
\endbibitem

\bibitem[\protect\citeauthoryear{Smulyan et~al.}{2016}]{jbhi2025_46}
\begin{barticle}
\bauthor{\bsnm{Smulyan}, \binits{H.}},
\bauthor{\bsnm{Lieber}, \binits{A.}},
\bauthor{\bsnm{Safar}, \binits{M.E.}}:
\batitle{Hypertension, diabetes type ii, and their association: role of arterial stiffness}.
\bjtitle{American journal of hypertension}
\bvolume{29}(\bissue{1}),
\bfpage{5}--\blpage{13}
(\byear{2016})
\end{barticle}
\endbibitem

\bibitem[\protect\citeauthoryear{Carr and Brunzell}{2004}]{jbhi2025_47}
\begin{barticle}
\bauthor{\bsnm{Carr}, \binits{M.C.}},
\bauthor{\bsnm{Brunzell}, \binits{J.D.}}:
\batitle{Abdominal obesity and dyslipidemia in the metabolic syndrome: importance of type 2 diabetes and familial combined hyperlipidemia in coronary artery disease risk}.
\bjtitle{The journal of clinical endocrinology \& metabolism}
\bvolume{89}(\bissue{6}),
\bfpage{2601}--\blpage{2607}
(\byear{2004})
\end{barticle}
\endbibitem

\bibitem[\protect\citeauthoryear{Luo et~al.}{2023}]{jbhi2025_48}
\begin{botherref}
\oauthor{\bsnm{Luo}, \binits{J.-Q.}},
\oauthor{\bsnm{Ren}, \binits{H.}},
\oauthor{\bsnm{Chen}, \binits{M.-Y.}},
\oauthor{\bsnm{Zhao}, \binits{Q.}},
\oauthor{\bsnm{Yang}, \binits{N.}},
\oauthor{\bsnm{Liu}, \binits{Q.}},
\oauthor{\bsnm{Gao}, \binits{Y.-C.}},
\oauthor{\bsnm{Zhou}, \binits{H.-H.}},
\oauthor{\bsnm{Huang}, \binits{W.-H.}},
\oauthor{\bsnm{Zhang}, \binits{W.}}:
Hydrochlorothiazide-induced glucose metabolism disorder is mediated by the gut microbiota via lps-tlr4-related macrophage polarization.
Iscience
\textbf{26}(7)
(2023)
\end{botherref}
\endbibitem

\end{thebibliography}

\end{document}